\documentclass[10pt,twocolumn,letterpaper]{article}

\usepackage{iccv}
\usepackage{times}
\usepackage{amsmath}
\usepackage{amssymb}
\usepackage[dvipdfmx]{color}
\usepackage[dvipdfmx]{graphicx} 

\newcommand{\figref}[1]{{Fig.~\ref{#1}}}
\newcommand{\tabref}[1]{{Tab.~\ref{#1}}}
\newcommand{\equref}[1]{{\eqref{#1}}} 

\newcommand{\refsec}[1]{Sec. \ref{sec:#1}}
\newcommand{\secref}[1]{\refsec{#1}} 

\newcommand{\refsubsec}[1]{Sec. \ref{subsec:#1}}

\newif\ifdraft
\drafttrue 

\newcommand{\first}[1]{{\color{red}{#1}}}
\newcommand{\second}[1]{{\color{blue}{#1}}}
\newcommand{\third}[1]{{\color{green}{#1}}}

\usepackage[pagebackref=true,breaklinks=true,letterpaper=true,colorlinks,bookmarks=false]{hyperref}

\def\iccvPaperID{4054} 

\iccvfinaltrue
\ificcvfinal\fi

\begin{document}

\title{TrTr: Visual Tracking with Transformer}

\author{Moju Zhao and Kei Okada and Masayuki Inaba\\
Department of Mechano-Infomatics, University of Tokyo\\
7-3-1 Hongo, Bunkyo-ku, Tokyo 113-8656, Japan\\
{\tt\small chou@jsk.imi.i.u-tokyo.ac.jp}
}

\maketitle
\ificcvfinal\thispagestyle{empty}\fi

\begin{abstract}
  Template-based discriminative trackers are currently the dominant tracking methods due to their robustness and accuracy, and the Siamese-network-based methods that depend on cross-correlation operation between features extracted from template and search images show the state-of-the-art tracking performance.
  However, general cross-correlation operation can only obtain relationship between local patches in two feature maps.
In this paper, we propose a novel tracker network based on a powerful attention mechanism called Transformer encoder-decoder architecture to gain  global and rich contextual interdependencies. 
In this new architecture, features of the template image is processed by a self-attention module in the encoder part to learn strong context information, which is then sent to the decoder part to compute cross-attention with the search image features processed by another self-attention module.
In addition, we design the classification and regression heads using the output of Transformer to localize target based on shape-agnostic anchor.
We extensively evaluate our tracker TrTr, on VOT2018, VOT2019, OTB-100, UAV, NfS, TrackingNet, and LaSOT benchmarks and our method performs favorably against state-of-the-art algorithms.
 Training code and pretrained models are available at \href{https://github.com/tongtybj/TrTr}{\color{green}{https://github.com/tongtybj/TrTr}}.
\end{abstract}

\section{Introduction}
\label{sec:introduction}

Visual object tracking is a task to estimate the state of a target object in a video sequence. Most commonly, the state is represented as a bounding box encapsulating the target in each frame, and the initial bounding box is given in the first frame. Although the tracking problem is closely related to the detection problem, the main difference from object detection is that detection only looks for a set of particular instances while tracking needs to even track unknown object class. Thus a tracker is required to extract proper features about the target from the initial frame (i.e., one-shot learning), and to localize the target in following frames. Among various tracking methods, template-based method is recently dominant due to their robustness on both classification and bounding box regression.


\begin{figure}[t]
  \begin{center}
    \includegraphics[width=\columnwidth]{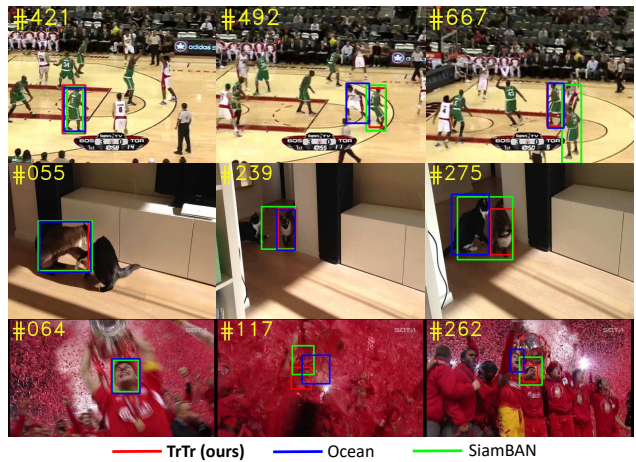}
    \caption{A Comparison of our proposed {\bf Tr}acker with {\bf Tr}ansformer (TrTr) with state-of-the-art trackers Ocean \cite{2020ECCV_Ocean} and SiamBAN \cite{2020CVPR_SiamBAN}. Unlike these trackers which depend on explicit cross-correlation operation between features extracted from the template and search images, our tracker applies Transformer encoder-decoder architecture \cite{2017NIPS_Transformer} which uses self- and cross-attention mechanisms to aggregate global and rich contextual interdependencies.  Observed from the visualization results, our tracer is more accurate, and robust to appearance changes, complex background and close distractors with occlusions.}
    \label{figure:abst}
  \end{center}
  \vspace{-5mm}
\end{figure}

The target classification is generally formulated as a confidence map upon the current search image. This has been achieved by the previously dominant Discriminative Correlation Filter (DCF) paradigms \cite{2010CVPR_MOSSE, 2015CVPR_CFLB, 2015TPAMI_KCF, 2016ECCV_C_COT, 2018CVPR_DRT} 
and the most recent Siamese-network-based trackers \cite{2016ECCVWS_SiamFC, 2017ICCV_DSiaM, 2018CVPR_RasNet, 2019ICCV_UpdateNet, 2020AAAI_SiamFC++}. Both methods exploit cross-correlation operation between features extracted from the template and search images to predict the appearance probability at each spatial position in search image for target localization.
For bounding box regression, recent Siamese-network-based trackers \cite{2018CVPR_SiamRPN, 2018ECCV_DaSiamRPN, 2019CVPR_SiamRPN++} introduce Region Proposal Network (RPN) which is common in detection problem, and develop a shared Siamese network for both classification and bounding box regression heads.
Given that the general cross-correlation operation can only obtain relationship between local patches in the template and search images, global pair-wise interaction cannot be performed, which makes it difficult to precisely track the target object with large appearance variations, close distractors,  or occlusions.
Another type of  recent trackers \cite{2019CVPR_ATOM, 2019ICCV_DiMP, 2020CVPR_PrDiMP} use IoUNet \cite{2018ECCV_IoUNet} and feature modulation vector to estimate bounding box globally in search image. However, the target classification branch of these trackers still depend on cross-correlation operation to train DCF.

Then, attention mechanisms for tracking are proposed by \cite{2018CVPR_FlowTrack, 2018CVPR_RasNet} to learn the global context from the whole image and enhance the discriminate ability between the target and close distractors or complex backgrounds. The most recent work SiamAttn \cite{2020CVPR_SiamAttn} adds self- and cross-attention modules for Siamese-network-based architecture, which outperforms its baseline structure SiamRPN++\cite{2019CVPR_SiamRPN++}. Alternatively, in this work, we introduce a new network architecture for tracking task, which exploits Transformer encoder-decoder architecture \cite{2017NIPS_Transformer} to perform both target classification and bounding box regression. In this new tracker, referred as TrTr, self-attention for both template and search image features, and cross-attention between these features are performed to compute global and rich contextual interdependencies, resulting in more accurate and stable tracking as shown in \figref{figure:abst}. 
In addition, we further equip our Transformer-based tracker with a plug-in online update module to capture the appearance changes of objects during tracking. This online update module further enhances the tracking performance, which shows the scalability of our proposed Transformer-based tracking approach.

The main contributions of this work are:
\begin{itemize}
\item We introduce Transformer encoder-decoder architecture for object vision tracking task, where explicit cross-correlation between feature maps extracted from the template and search images is replaced  with the self- and cross-attention operations to gain global and rich contextual interdependencies.
\item The confidence based target classification head and shape-agnostic anchor based target regression head are developed for our Transformed-based architecture; while a plug-in online update module for classification is designed to further enhances the tracking performance.

\item We conduct comprehensive experiments on large scale benchmark datasets including VOT, OTB, UAV, NfS, LaSOT, and TrackingNet, and our tracker achieves favorable performance compared with the state-of-the-art results while running in real-time speeds, which demonstrates a significant potential of Transformer architecture in tracking task.
\end{itemize}

\section{Related Work}
\label{sec:related_work}

Visual tracking is one of the most active research topics in computer vision in recent decades. A comprehensive survey of the related trackers is beyond the scope of this paper, so we only briefly review two aspects that are most relevant to our work: the templated-based visual object tracking and the Transformer architecture.

\subsection{templated-based visual object tracking}
Template-based trackers depend on a target template given in the initial frame. The most recent state-of-the-art tracking methods can be roughly divided into two categories: correlation-based filter based trackers and Siamese-network-based trackers.

The correlation-based trackers \cite{2015TPAMI_KCF, 2015ICCV_HCFCF, 2018ECCV_UPDT}  rely on diagonalizing transformation of circular convolutions, given by the Discrete Fourier Transform, to perform efficient online training on correlation filter. Thus, the target template can be updated online to improve the discriminative robustness. Although these trackers can provide the location of the target from confidence map on search image, the  target shape is difficult to be estimated by such trackers. Commonly, a multi-scale strategy is applied to handle the change in target size, which however cannot address shape deformation. Most recent related works \cite{2019CVPR_ATOM, 2019ICCV_DiMP, 2020CVPR_PrDiMP} exploit deep learned features from backbone based on convolutional neural networks (CNN) for not only correlation-based classification but also target regression using  another network (e.g., IoUNet \cite{2018ECCV_IoUNet}). Thus, such CNN-based trackers have two independent components: online trained filter for target classification and offline trained network for target regression.

The Siamese-network-based tracker, which is first proposed by  SiamFC \cite{2016ECCVWS_SiamFC}, extracts deep features from the template and the search images using the same offline-trained CNN backbone, then performs cross-correlation between these features to compute the matching scores for target localization. Further, SiamRPN \cite{2018CVPR_SiamRPN} borrows the RPN idea from object detectors to enable bounding box regression, which achieves an end-to-end learning tracker for both classification and regression.
Then, SiamRPN++\cite{2019CVPR_SiamRPN++} is developed to aggregate multiple features from different layers of backbone to perform layer-wise cross-correlation, which serves as the baseline structure for recent state-of-the-art Siamese-network-based trackers \cite{2020CVPR_SiamCAR, 2020CVPR_SiamBAN, 2020CVPR_SiamAttn, 2020CVPR_LTMU}.

The proposed network in our work has similar network structure with SiamRPN++, which also contains a shared backbone for feature extraction and heads for classification and regression. However, instead of applying cross-correlation layer which is the core of Siamese network, we use Transformer encoder-decoder architecture and its attention mechanism.

\subsection{Transformer encoder-decoder architecture}
Transformer \cite{2017NIPS_Transformer} is developed as a new attention-based building block for machine translation. Attention mechanisms \cite{2015NIPS_Attention}  are neural network layers that aggregate information from the entire input sequence.
Transformer introduces self-attention layers to scan through each element of a sequence and update it by aggregating information from the whole sequence. Such a self-attention is performed for both the input of encoder and decoder. Then, cross-attention (encoder-decoder attention) layer computes the interdependency between two sequences.

Transformer is now replacing Recurrent Neural Network (RNN) in many problems in natural language processing and speech processing \cite{2019NAACL_BERT, 2019INTERSPEECH_RWTH, 2019_GPT, 2019ArXiV_ASR}.
The original Transformer \cite{2017NIPS_Transformer} is first used in auto-regressive models, following early sequence-to-sequence models \cite{2014NIPS_Seq2Seq} and generating output tokens one by one. Then, parallel sequence generation is developed in the domains of machine translation and audio \cite{2018ICML_ParalleWaveNet, 2019arXiv_MaskPredict}  to address the prohibitive inference cost (proportional to output length, and hard to batch).
In computer vision, Transformer with parallel decoding is being introduced in image classification \cite{2020arXiv_ViT} and object detection \cite{2020ECCV_DETR, 2020arXiv_DeformableDETR}, which demonstrate state-of-the-art results in both fields. Further, Transformer is exploited in multiple object tracking task \cite{2021arXiv_TrackFormer, 2020arXiv_TransTrack}, of which however the target classes are pre-defined. Therefore, in this work, we develop a new tracker based on Transformer architecture and its attention mechanism to enable tracking class-agnostic target, which is the main challenge of single object tracking task.

\section{Transformer Architecture}
\label{sec:transformer}

\begin{figure}[h]
  \begin{center}
    \includegraphics[width=\columnwidth]{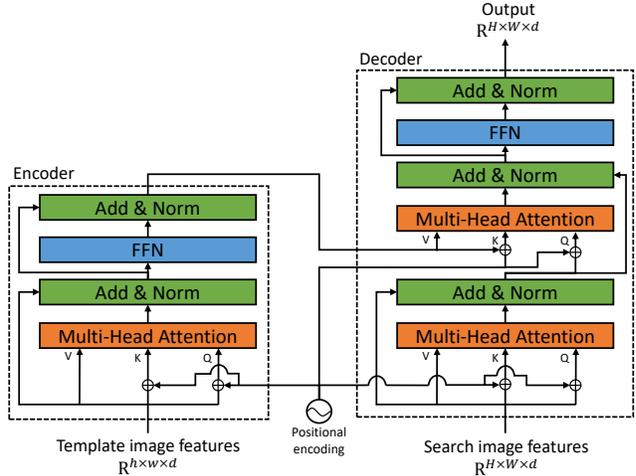}
    \caption{Architecture of TrTr Transformer for tracking task, which replaces cross-correlation operation in Siamese network with multi-head self- and encoder-decoder attention mechanisms.}
    \label{figure:transformer}
  \end{center}
  \vspace{-5mm}
\end{figure}

This section presents the Transformer encoder-decoder architecture for visual tracking. As shown in \figref{figure:transformer}, the input features to Transformer are extracted from a shared backbone network elaborated later in \refsubsec{backbone}, while the output is used for classification and regression heads presented in \refsubsec{branches}.
Our Transformer architecture follows the original structure in \cite{2017NIPS_Transformer}, which consists of two components: encoder with template image features, and decoder with search image features. There are multi-head attention modules in both encoder and decoder, which are the key to perform self-attention with a feature sequence and cross-attention between two feature sequences.

\subsection{Encoder and Decoder Components}

Unlike the original Transformer \cite{2017NIPS_Transformer}, our Transformer model only contains one layer for both encoder and decoder.

\paragraph{Encoder.}
The encoder has a standard architecture and consists of a multi-head self-attention module and a feed forward network (FFN) which are elaborated later.
We assume there is a template image feature map $z_0 \in \mathcal{R}^{h\times w \times d}$, where  $h$ and $w$ are the height and width of the feature map, and $d$ is channel dimension.  Then, we collapse the spatial dimensions of $z_0$ into one dimension  as a sequence with the size of $\mathcal{R}^{hw \times d}$.
Since the transformer architecture is permutation-invariant, we add fixed positional encodings \cite{2017NIPS_Transformer} for this input sequence.

\paragraph{Decoder.}
The decoder also follows the standard architecture, which contains multi-headed attention modules for self- and encoder-decoder (cross-) attention followed by a FFN.
The input of decoder is a search feature map $x_0 \in \mathcal{R}^{H \times W \times d}$, of which the channel dimension is same with $z_0$, but the spatial dimensions are larger ($W > w, H >h$) for tracking purpose. Similar to the encoder, the spatial dimensions of $x_0$ is also collapsed into one dimension, resulting in a sequence of $\mathcal{R}^{HW \times d}$. The fixed positional encodings are also added in this sequence. Following the strategy in \cite{2020ECCV_DETR}, our model decodes all elements in this sequence in parallel. The output has the same size as input ($\mathcal{R}^{HW \times d}$), which is finally reverted to $\mathcal{R}^{H \times W \times d}$ for subsequent classification and regression.


\subsection{Attention Mechanism}

Following \cite{2017NIPS_Transformer},  query-key-value attention mechanism is applied. Query, key, and value sequences, $Q \in \mathcal{R}^{N_\mathrm{q} \times d^{'}}$, $K  \in \mathcal{R}^{N_\mathrm{kv}  \times d^{'}}$, and $V  \in \mathcal{R}^{N_\mathrm{kv}  \times d^{'}} $ are obtained by linear projection with weight matrices $W_{\mathrm{q}}, W_{\mathrm{k}}, W_{\mathrm{v}} \in \mathcal{R}^{d\times d^{'}}$:
\begin{equation}
  \label{eq:qkv_matrices}
  \left[Q; K; V\right] = \left[ (X_{\mathrm{q}} + P_{\mathrm{q}})W_{\mathrm{q}}); (X_{\mathrm{kv}} + P_{\mathrm{k}})W_{\mathrm{k}}); X_{\mathrm{kv}}W_{\mathrm{v}}) \right],
\end{equation}
where $X_{\mathrm{q}} \in \mathrm{R}^{N_{\mathrm{q}} \times d}$, $ X_{\mathrm{kv}} \in  \mathrm{R}^{N_{\mathrm{kv}} \times d}$ are spatially one-dimensional inputs, respectively. For self-attention module in encoder, $N_{\mathrm{q}} = N_{\mathrm{kv}} = hw$, and $X_{\mathrm{q}} = X_{\mathrm{kv}}$; for self-attention module in decoder, $N_{\mathrm{q}} = N_{\mathrm{kv}} = HW$, and $X_{\mathrm{q}} = X_{\mathrm{kv}}$. For encoder-decoder attention module, $N_{\mathrm{q}} = HW$,  $N_{\mathrm{kv}} = hw$. $P_{\mathrm{q}}$ and $P_{\mathrm{k}}$ are the positional encodings for query and key sequences, respectively; however we do not add positional encodings for value sequence $V$ similarly to  \cite{2020ECCV_DETR}.

The attention weight map $A \in \mathrm{R}^{N_{\mathrm{q}} \times N_{\mathrm{kv}}} $ is then computed based on the softmax of dot products between query and key sequences to obtain the pair-wise correlation between them:
\begin{eqnarray}
  \label{eq:qk_weights}
  A_{ij} = \frac{e^{\frac{1}{\sqrt{d^{'}}} Q_i^{\mathrm{T}} K_j} }{Z_i} \hspace{3mm} \mathrm{where} \hspace{1mm} Z_{i} = {\displaystyle \sum_{j = 1}^{N_{\mathrm{kv}}}} e^{\frac{1}{\sqrt{d^{'}}} Q_i^{\mathrm{T}} K_j },
\end{eqnarray}
where $Q_i, K_j \in \mathrm{R}^{d^{'}}$ are the $i$-th and $j$-th vector of $Q$ and $K$. From the view of image processing, $A_{i,j}$ represents the correlation value between the position $i$ in query image and the position $j$ in key image.
The final output of attention is the aggregation of values weighted by attention weight $A$:
\begin{eqnarray}
  \label{eq:attention_output}
  Attn(X_{\mathrm{q}}, X_{\mathrm{kv}}, W) = A V,
\end{eqnarray}
where $W$ is the concatenation of $W_{\mathrm{q}}, W_{\mathrm{k}}$, and $W_{\mathrm{v}}$.

The multi-head attention is simply the concatenation of $M$ single attention heads followed by a linear projection:
\begin{eqnarray}
  \label{eq:multihead_attn}
&  MultiHeadAttn(X_{\mathrm{q}}, X_{\mathrm{kv}}) \nonumber \\
&  = \left[Attn(X_{\mathrm{q}}, X_{\mathrm{kv}}, W_1);\cdots; Attn(X_{\mathrm{q}}, X_{\mathrm{kv}}, W_M)\right] W^{o}.
\end{eqnarray}
The attention outputs are concatenated on channel axis, and thus in each single attention head, the channel dimension is $d^{'} = \frac{d}{M}$; whereas,  $W^{o} \in \mathrm{R}^{d\times d}$.

Then, a residual connections and layer normalization is further used according to the common practice in \cite{2017NIPS_Transformer}:
\begin{eqnarray}
  \label{eq:residual_layer_norm}
X_{\mathrm{q}}^{'} = \mathrm{layernorm}(MultiHeadAttn(X_{\mathrm{q}}, X_{\mathrm{kv}}) + X_{\mathrm{q}}).
\end{eqnarray}

After each self- and encoder-decoder attention module, a feed-forward network (FFN) composed of two-layers of 1x1 convolutions with ReLU activations is used to process $X_{\mathrm{q}}^{'}$. The dimension of input and output channels are $d$, while the hidden layer has a larger channel dimension. There is also a residual connection and layernorm after the FFN, similarly to \cite{2017NIPS_Transformer}.

Using these self-attention and encoder-decoder attention mechanisms over inputs, the model can globally reason about input features together using pair-wise relations between them, which helps to discriminate between foreground and background.
In our model, we use 8 heads for multi-head attention module ($M=8$), and set the channel dimension of FFN hidden layer to $8d$.

\section{Tracker with Transformer}
\label{sec:network}
\begin{figure*}[h]
  \begin{center}
    \includegraphics[width=2\columnwidth]{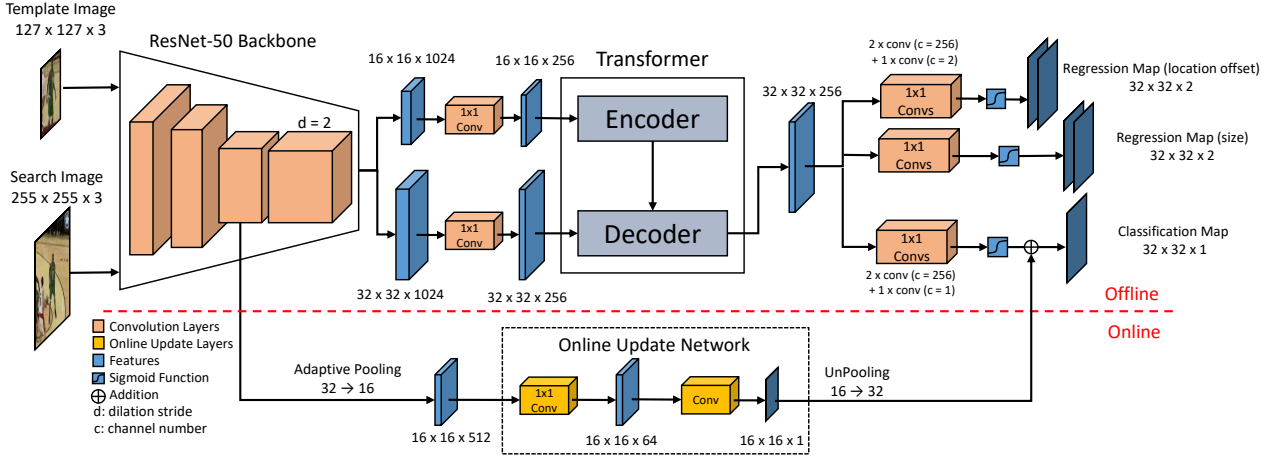}
    \caption{Overview of the proposed tracking framework, consisting of an offline tracking part (top) and an online classification model update part (bottom). The offline tracking part includes a backbone for feature extraction (\refsubsec{backbone}), Transformer module (\secref{transformer}) and heads for classification and regression (\refsubsec{branches}). The plug-in online update network models the appearance changes of target object to further improve the classification accuracy, as detailed in \refsubsec{online}. The size of target template image is $127\times 127$, while the size of search image is configurable, and its typical value is $255 \times 255$.}
    \label{figure:network}
  \end{center}
\end{figure*}

This section depicts the tracking algorithm building upon the proposed Transformer architecture. It contains two parts: an offline model based on Transformer and an online update model only for classification, as illustrated in \figref{figure:network}.

\subsection{Feature Extraction}
\label{subsec:backbone}
Our approach takes a pair of images as inputs, i.e., a template image and a candidate search image. The template image represents the object of interest, i.e., an image patch centered on the target object in the initial frame. The search image is typically larger and represents the search area in subsequent video frames. Feature extraction for template and search images share a modified ResNet-50 \cite{2016CVPR_RESNET} backbone. The last stage of the standard ResNet-50 is truncated, thereby only the first fourth stages are used. In the fourth stage, the convolution stride of down-sampling unit is modified from 2 to 1 to increase the spatial size of feature maps, meanwhile, all the $3 \times 3$ convolutions are augmented with a dilation with stride of 2 to increase the receptive fields. These modifications increase the resolution of output features, thus improving the feature capability on object localization \cite{2018TPAMI_DeepLab, 2019CVPR_SiamRPN++}. The features from backbone are then passed through a 1 $\times$ 1 convolution layer to reduce the channel dimension from 1024 to 256 for saving computation cost in subsequent Transformer module. 

\subsection{Target Localization}
\label{subsec:branches}
To localize target object and estimate the shape, we apply the shape-agnostic anchor \cite{2019arXiv_CenterNet}. Three independent heads are connected to the output of Transformer module. One is for target classification and the other two are for regression. Each head contains three $1 \times 1$ convolution layers followed by a Sigmoid layer. The classification head yields a map $Y \in [0, 1]^{\lfloor \frac{H}{s} \rfloor \times \lfloor \frac{W}{s} \rfloor \times 1}$, where $H, W$ are the height and the width of search image (i.e., both are 255 in \figref{figure:network}), and $s$ is the output stride (i.e., 8). $\lfloor  \rfloor$ is the floor function, which leads to a result of 32 in \figref{figure:network}. The map $Y$ corresponds to the appearance probability of the target in discretized low resolution. Therefore, to recover the discretization error caused by the output stride $s$, it is necessary to  additionally predict a local offset $O \in [0, 1)^{\lfloor \frac{H}{s} \rfloor \times \lfloor \frac{W}{s} \rfloor \times 2}$. Then the center point of the target object in search image can be given by:
\begin{eqnarray}
  \label{eq:target_center}
  (x_c, y_c) = s(argmax(Y^{'}) + O(argmax(Y^{'}))),
\end{eqnarray}
where $Y^{'}$ is combination of the raw map $Y$ with a cosine window to suppress the large displacement, similarly to \cite{2018CVPR_SiamRPN, 2019CVPR_SiamRPN++}. $argmax(Y^{'})$ returns a 2D location corresponding to the peak of map $Y{'}$.

For size regression, we design another head to yield a normalized map $S \in [0, 1]^{\lfloor \frac{H}{s} \rfloor \times \lfloor \frac{W}{s} \rfloor \times 2}$. Then, the size of bounding box in search image can be given by:
\begin{eqnarray}
  \label{eq:target_size}
  (w_{bb}, h_{bb}) = (W, H) * S(argmax(Y^{'})),
\end{eqnarray}
where $*$ is  element-wise multiplication operation. Linear interpolation update strategy \cite{2018CVPR_SiamRPN, 2019CVPR_SiamRPN++} is further applied to smooth the change in bounding box size.
Finally, corners of bounding box can be easily calculated from the target center point $(x_c, y_c)$ and the size $(w_{bb},h_{bb})$.

\subsection{Loss Function}
We follow the loss function for object detection proposed by \cite{2019arXiv_CenterNet}; however, there is only a single class to classify in our network, and the bounding box is normalized. For a ground truth target center $\bar{p}$, we first compute a low-resolution equivalent $\tilde{p} = (\lfloor \frac{\bar{p}_x}{s} \rfloor, \lfloor \frac{\bar{p}_y}{s} \rfloor)$. Then we use a Gaussian kernel $\bar{Y} = \mathrm{exp}(- \frac{(x - \tilde{p}_x)^2 + (y - \tilde{p}_y)^2}{2 \sigma^2_p})$ to compare with the predicted map $Y$, where $\sigma_p$ is an object size-adaptive standard deviation \cite{2018ECCV_CornerNet}. The training objective for classification is a penalty-reduced pixel-wise logistic regression with focal loss \cite{2018ICCV_FocalLoss}:
\begin{eqnarray}
  \label{eq:classification_loss}
  L_Y = - \sum_{xy}
  \begin{cases}
    (1 - Y_{xy})^{\alpha}\mathrm{log}(Y_{xy}) & (\bar{Y}_{xy} = 1) \\ \\
    \begin{split}
      & (1 - \bar{Y}_{xy})^{\beta} (Y_{xy})^{\alpha} \\
      & \mathrm{log}(1 - Y_{xy})
    \end{split} & (otherwise)
  \end{cases}
\end{eqnarray}
where $Y_{xy}$ is the value of map $Y$ at $(x, y)$. $\alpha$ and $\beta$ are hyper-parameters of the focal loss \cite{2018ICCV_FocalLoss}. We use $\alpha = 2$ and $\beta = 4$ in all our experiments, following \cite{2018ECCV_CornerNet}.

Then, the loss function for offset regression is formulated using L1 loss:
\begin{eqnarray}
  \label{eq:offset_loss}
  L_O = |O_{\tilde{p}} - (\frac{p}{s} - \tilde{p}) |,
\end{eqnarray}
where, $O_{\tilde{p}}$ is the map value of $O$ at $\tilde{p}$.

For a ground truth bounding box size ($\bar{w}_{bb}$, $\bar{h}_{bb}$), we also use L1 loss similar to \equref{eq:offset_loss}:
\begin{eqnarray}
  \label{eq:offset_loss}
  L_S = |S_{\tilde{p}} - \tilde{s} |,
\end{eqnarray}
where, $S_{\tilde{p}}$ is the map value of $S$ at $\tilde{p}$, and $\tilde{s} = (\frac{\bar{w}_{bb}}{W}, \frac{\bar{h}_{bb}}{H})$ is normalized ground truth bounding box size.
Finally, the joint training objective for the entire network can be given by:
\begin{eqnarray}
  \label{eq:joint_loss_function}
  L = L_Y + \lambda_1 L_O + \lambda_2 L_S,
\end{eqnarray}
where $\lambda_1$ and $\lambda_2$ are the trade-off hyper-parameters. We do not search for these hyper-parameters, and simply set $\lambda_1 = \lambda_2 = 1$.

\subsection{Integrating Online Update for Classification}
\label{subsec:online}
An online update model inspired by \cite{2019CVPR_ATOM, 2019ICCV_DiMP} is designed to independently capture the appearance changes of target object during tracking. As shown in the bottom part of \figref{figure:network}, this online branch directly uses the output from the first third stage of the backbone network and generates a map $Y_{online} \in [0, 1]^{\lfloor \frac{H}{s} \rfloor \times \lfloor \frac{W}{s} \rfloor \times 1}$. This branch consists of 2-layer fully convolutional neural network, where the first $1 \times 1$ convolutional layer reduces the channel dimension to 64, and the second layer employs $4 \times 4$ kernel with a single output channel. The fast conjugate gradient algorithm proposed in \cite{2019CVPR_ATOM} is applied to train this online network during inference.
The maps estimated by the offline classification head and the online branch  are weighted as
\begin{eqnarray}
  \label{eq:online_module}
Y^{''} = w Y^{'} + (1 - w) Y_{online},
\end{eqnarray}
where $w$ denotes the trade-off hyper-parameter, which is set to 0.6 in our experiments.
When the online update model is available, the combined classification map $Y^{''}$ is utilized instead of $Y^{'}$ in \equref{eq:target_center} and \equref{eq:target_size}. We refer readers to \cite{2019CVPR_ATOM, 2019ICCV_DiMP} for more details on the online update model.


\section{Experiments}
\label{sec:experiments}

\subsection{Implementation Details}
By following SiamFC \cite{2016ECCVWS_SiamFC}, we adopt a template image of 127 $\times$ 127 and a search image of 255 $\times$ 255 for training. 
The backbone network ResNet-50 is initialized with the parameters pretrained on ImageNet \cite{2015IJCV_ImageNet}. The whole network  is then trained on the datasets of Youtube-BB \cite{2017CVPR_YouTubeBB}, ImageNet VID \cite{2015IJCV_ImageNet}, GOT-10k \cite{2019TPAMI_GOT10k}, LaSOT \cite{2019CVPR_LaSOT}, and COCO \cite{2014ECCV_COCO}, which yield  $4.7\times 10^5$ training pairs for one epoch.
We apply ADAM \cite{2015ICLR_Adam} optimization on 8 GPUs, with each GPU hosting 64 image pairs, and thus the mini-batch size is 512 image pairs per iteration.
There are 20 epochs in total with learning rates starting from $10^{-5}$ for backbone part and $10^{-4}$ for other parts, and these learning rates are decayed with 0.5 every 5 epochs. Our approach is implemented in Python using PyTorch on a PC with Intel(R) Xeon(R) W-3225 CPU 3.70GHz, 64G RAM, Nvidia RTX 3090.



During inference, the size of search image is configurable, and  we use the default size of 255 $\times$ 255 in most cases. In case of fast target motion, we choose a larger size (e.g., 280 $\times$ 280 or 320 $\times$ 320) to ensure the appearance of the target in the next search image.
The tracking speed of the offline model (top part of \figref{figure:network}) with a search image of 255 $\times$ 255 is 50 FPS in our experiment environment, and will reduce to 35 FPS if integrating online update model (the bottom part of \figref{figure:network}).



\subsection{Comparison with the state-of-the-art}

We compare our approach, termed TrTr, on seven tracking benchmarks. We evaluate two versions: TrTr-offline for only the offline part shown in \figref{figure:network}, and TrTr-online for combination with plug-in online update module.

{\bf On VOT2018 \& VOT2019.}
We evaluate on the 2018 edition of the Visual Object Tracking challenge \cite{2018VOT} by comparing with the state-of-the-art methods.
The dataset contains 60 videos, and trackers are restarted at failure by the evaluation protocol. The performance is then decomposed into accuracy and robustness,
defined using IoU overlap and failure rate respectively. The main EAO metric takes both these aspects into account. The size of search image in our tracker is set to 280 $\times$ 280. The results, computed over 15 repetitions as specified in the protocol, are shown in \tabref{table:vot}. Our offline TrTr tracker outperforms the representative Siamese-network based tracker SiamRPN++ \cite{2019CVPR_SiamRPN++} by 1.0 point in terms of EAO. It is worth noting that the improvements mainly come from the accuracy score, which obtains 1.2 points relative increases over SiamRPN++.
Our online augmented model further improves our tracker by 6.9 points in terms of EAO, which achieves the best overall performance with the highest robustness and competitive accuracy compared to previous methods.

VOT2019 \cite{2019VOT} is refreshed by replacing the 12 least difficult sequences from the previous version with several carefully selected sequences in GOT-10k dataset \cite{2019TPAMI_GOT10k}. The same measurements are exploited for performance evaluation; however, we expand the search image size to 320 $\times$ 320 to handle fast target motion in several challenge sequences. As illustrated in \tabref{table:vot}, our offline model marks the best accuracy score of 0.608; while our online augmented model achieves an accuracy of 0.601, a robustness of 0.228 and an EAO of 0.384, which ranks second best and also outperforms DRNet \cite{2019VOT} (leading performance on the public dataset of VOT2019 challenge) in terms of robustness. 

\begin{table}[!t]
  \begin{center}
    \begin{tabular*}{\columnwidth}{@{\extracolsep{\fill}}l @{\extracolsep{\fill}} c @{\extracolsep{\fill}} c @{\extracolsep{\fill}}c @{\extracolsep{\fill}}c @{\extracolsep{\fill}}c @{\extracolsep{\fill}}c @{\extracolsep{\fill}}}
      \hline
       & \multicolumn{3}{c}{VOT2018} & \multicolumn{3}{c}{VOT2019} \\ \cline{2-7}
      & A $\uparrow$ & R $\downarrow$ & EAO $\uparrow$ & A $\uparrow$ & R $\downarrow$ & EAO $\uparrow$ \\ \cline{2-7} \hline
      \small{ATOM \cite{2019CVPR_ATOM}} & \small{0.590} & \small{0.204} & \small{0.401}  & \small{\third{0.603}} & \small{0.411}  & \small{0.292}  \\
      \small{SiamRPN++ \cite{2019CVPR_SiamRPN++}} & \small{0.600} & \small{0.234} & \small{0.414}  & \small{0.599}  & \small{0.482}  & \small{0.285}  \\
      \small{SiamFC++ \cite{2020AAAI_SiamFC++}} & \small{0.587} & \small{0.183} & \small{0.426}  & -  & -  & -  \\
      \small{DiMP50 \cite{2019ICCV_DiMP}} & \small{0.597} & \small{0.153} & \small{0.440}  & \small{0.594}  & \small{\third{0.278}}  & \third{\small{0.379}}  \\
      \small{PrDiMP50 \cite{2020CVPR_PrDiMP}} & \third{\small{0.618}} & \small{0.165} & \small{0.442}  & -  & -  & -  \\
      \small{SiamBAN \cite{2020CVPR_SiamBAN}} & \small{0.597} & \small{0.178} & \small{0.452}  & \small{0.602}  & \small{0.396}  & \small{0.327} \\
      \small{MAML-Retina \cite{2020CVPR_MAML}} & \small{0.604} & \small{0.159} & \small{0.452}  & \small{0.570} & \small{0.366} & \small{0.313}  \\
      \small{SiamAttn \cite{2020CVPR_SiamAttn}} & \second{\small{0.630}} & \small{0.160} & \third{\small{0.470}}  & -  & -  & -  \\
      \small{Ocean \cite{2020ECCV_Ocean}} & \small{0.592} & \small{\second{0.117}} & \small{\second{0.489}}  & \small{0.594} & \small{0.316} & \small{0.350}  \\
      \small{DRNet \cite{2019VOT}} & - & - & -  & \small{\second{0.605}}  & \small{\second{0.261}}  & \small{\first{0.395}}  \\
      \small{D3S \cite{2020CVPR_D3S}} & \small{\first{0.640}} & \third{\small{0.150}} & \small{\second{0.489}}  & -  & -  & -  \\ \hline
      \small{TrTr-offline}  & \small{0.612} & \small{0.234} & \small{0.424}  & \first{\small{0.608}}  & \small{0.441}  & \small{0.313}  \\
      \small{TrTr-online}   & \small{0.606} & \small{\first{0.110}} & \small{\first{0.493}}  & \small{0.601}  & \first{\small{0.228}}  & \small{\second{0.384}}  \\ \hline
    \end{tabular*}
    \caption{Comparison with SOTA trackers on VOT2018 and VOT2019, with accuracy (A), robustness (R), and expected average overlap (EAO). \first{Red}, \second{blue} and \third{green} fonts indicate the top-3 trackers. ``TrTr'' denotes our proposed model.}
    \label{table:vot}
  \end{center}
\end{table}

\begin{figure}[t]
  \begin{center}
    \includegraphics[width=\columnwidth]{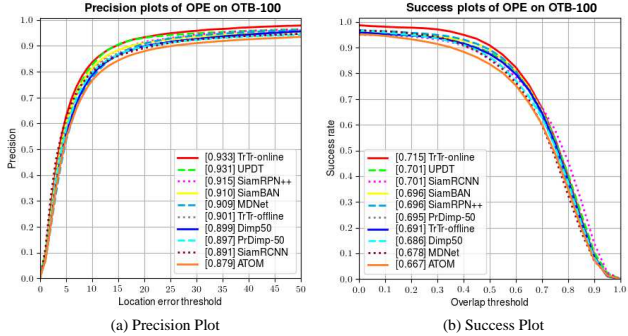}
    \caption{Comparison with state-of-the-art methods on success and precision plots on OTB-100.}
    \label{figure:otb}
  \end{center}
\end{figure}

{\bf On OTB-100.} We evaluate both our trackers TrTr-offline and TrTr-online on OTB-100 benchmark \cite{2015TPAMI_OTB}, which contains 100 sequences with 11 annotated video attributes.
The size of search image is set to 280 $\times$ 280.
\figref{figure:otb} compares our trackers with some recent top-performing trackers.
We follow the one pass evaluation (OPE) protocol, and report the AUC scores of success plot.
TrTr-offline and TrTr-online obtain success AUC scores  of 0.691 and 0.715, respectively. To the best of our knowledge, TrTr-online is the best-performing tracker ever on OTB-100.


{\bf On UAV123.}
This challenging dataset \cite{2016ECCV_UAV123}, containing 123 videos, is designed to benchmark trackers for UAV applications. It features small objects, fast motions, occlusion, absent, and distractor objects.
The size of search image in our tracker is set to 350 $\times$ 350 to handle fast moving target.
We show the result in \tabref{table:nfs_uav}. Our online augmented model obtain 65.2\% in terms of overall AUC score, which ranks third place, and outperforms the state-of-the-art Siamese-network-based trackers SiamRPN++\cite{2019CVPR_SiamRPN++} and  Siam R-CNN \cite{2020CVPR_SiamRCNN} by 1 point and 0.3 point, respectively.

{\bf On NfS.}
We evaluate our approach on need for speed dataset \cite{2017ICCV_NFS} (30 FPS version), which is the first high frame rate dataset recorded at real-world scenes. It includes 100 fully annotated videos (380K frames) with fast moving target objects, and we set the size of search image in our tracker to 380 $\times$ 380 to track the fast moving targets.
The AUC  scores of state-of-the-art trackers are shown in \tabref{table:nfs_uav}. Our online augmented model marks the third best.

\begin{table}[!t]
  \begin{center}
    \begin{tabular*}{\columnwidth}{@{\extracolsep{\fill}}l @{\extracolsep{\fill}} c @{\extracolsep{\fill}} c @{\extracolsep{\fill}}c @{\extracolsep{\fill}}c @{\extracolsep{\fill}} c @{\extracolsep{\fill}} c @{\extracolsep{\fill}} c @{\extracolsep{\fill}} c @{\extracolsep{\fill}} c @{\extracolsep{\fill}} }
      \hline
      & \shortstack{\\ \tiny{MDNet} \\ \scriptsize{\cite{2016CVPR_MDNet}} } & \shortstack{  \tiny{SiamRPN++} \\ \scriptsize{\cite{2019CVPR_SiamRPN++}} } & \shortstack{  \tiny{ATOM} \\ \scriptsize{\cite{2019CVPR_ATOM}} } & \shortstack{  \tiny{DiMP50} \\ \scriptsize{\cite{2019ICCV_DiMP}} } & \shortstack{  \tiny{SiamBAN} \\ \scriptsize{\cite{2020CVPR_SiamBAN}} } & \shortstack{  \tiny{SiamRCNN} \\ \scriptsize{\cite{2020CVPR_SiamRCNN}} } & \shortstack{  \tiny{PrDiMP50} \\ \scriptsize{\cite{2020CVPR_PrDiMP}} } & \shortstack{ \scriptsize{TrTr} \\ \scriptsize{offline} } & \shortstack{  \scriptsize{TrTr} \\ \scriptsize{online}} \\ \hline
      \scriptsize{UAV123} & \scriptsize{52.8}  & \scriptsize{61.3} &\scriptsize{65.0} &\scriptsize{\second{65.3}} &\scriptsize{63.1} &\scriptsize{64.9} & \scriptsize{\first{68.0}} & \scriptsize{59.4} & \scriptsize{\third{65.2}} \\ \hline
      \scriptsize{NfS} & \scriptsize{42.2} & \scriptsize{-} & \scriptsize{59.0}  &\scriptsize{61.9} &\scriptsize{59.4} &\scriptsize{\first{63.9}}  & \scriptsize{\second{63.5}} & \scriptsize{55.2} & \scriptsize{\third{63.1}}\\ \hline
    \end{tabular*}
    \caption{Comparison with SOTA trackers on UAV123 and  NfS datasets, with AUC. \first{Red}, \second{blue} and \third{green} fonts indicate the top-3 trackers. ``TrTr'' denotes our proposed model.}
    \label{table:nfs_uav}
  \end{center}
\end{table}

{\bf On TrackingNet.}
This is a large-scale tracking dataset with high variety in terms of classes and scenarios \cite{2018ECCV_TrackingNet}, and the test set contains over 500 videos without publicly available ground-truth.
The size of search image in our tracker is set to 320 $\times$ 320.
The results, shown in \tabref{table:trackingnet_lasot}, are obtained through an online evaluation server.
Our online augmented model has a comparable score in terms of the normalized precision (N-prec.); however, AUC is relatively lower than the state-of-the-art methods.
We observe that when the target object is considerably large and close to the whole image size (which are often seen in TrackingNet), the predicted bounding box from our tracker is always smaller than the ground truth, which significantly affects the success rate. This is the reason why our model has a higher N-prec. than SiamRPN++\cite{2019CVPR_SiamRPN++} and DiMP50 \cite{2019ICCV_DiMP}, but AUC is lower than these methods.

{\bf On LaSOT.}
Last, we evaluate on the large-scale LaSOT dataset \cite{2019CVPR_LaSOT}. The test set contains 280 long videos (2500 frames in average), thus emphasizing the robustness of the tracker, along with its accuracy.
The size of search image in our tracker is set to 350 $\times$ 350 to handle fast moving target.
As shown in \tabref{table:trackingnet_lasot}, our trackers show a relative loss on this benchmark compared with the state-of-the-art results.
We suspect that our tracker model is not robust enough to handle the very long sequences where the target object is always occluded or absent.

\begin{table}[!t]
  \begin{center}
    \begin{tabular*}{\columnwidth}{@{\extracolsep{\fill}}l @{\extracolsep{\fill}} c @{\extracolsep{\fill}} c @{\extracolsep{\fill}}c @{\extracolsep{\fill}}}
      \hline
       & \multicolumn{2}{c}{TrackingNet} & LaSOT \\ \cline{2-4}
      & AUC (\%) & N-Prec. (\%) &  AUC (\%) \\  \cline{2-4} \hline
      ATOM \cite{2019CVPR_ATOM} & {70.3} & {77.1} & {51.5} \\
      SiamRPN++ \cite{2019CVPR_SiamRPN++} & {73.3} & {80.0} & {49.6} \\
      DiMP50 \cite{2019ICCV_DiMP} & {74.0} & {80.1} & \third{56.8} \\
      SiamAttn \cite{2020CVPR_SiamAttn} & {75.2} & \third{81.7} & {56.0}  \\
      MAML-Retina \cite{2020CVPR_MAML} & \third{75.7} & \second{82.2} & {52.3} \\
      PrDiMP50 \cite{2020CVPR_PrDiMP} & \second{75.8} & {81.6} & \second{59.8} \\
      SiamRCNN \cite{2020CVPR_SiamRCNN} & \first{81.2} & \first{85.4}  & \first{64.8} \\\hline
      TrTr-offline  & {69.3} & {77.2} & {46.3} \\
      TrTr-online   & {71.0}  & {80.3} & {55.1} \\ \hline
    \end{tabular*}
    \caption{ Comparison with SOTA trackers on large scale dataset  TrackingNet and LaSOT, with AUC and norm precision (N-Prec.). \first{Red}, \second{blue} and \third{green} fonts indicate the top-3 trackers. ``TrTr'' denotes our proposed model.}
    \label{table:trackingnet_lasot}
  \end{center}
\end{table}

\subsection{Ablation Study}

Here, we perform an extensive analysis of the proposed networks by conducting ablation study on VOT2018.

\paragraph{Number of Transformer layers.}
The number of encoder and decoder layers significantly influences the performance of Transformer as reported in \cite{2017NIPS_Transformer}. We train models with different combination of encoder and decoder layers (up to 3 layers for both encoder and decoder), and compare the tracking performance as shown in \tabref{table:transformer_layers}.
It is very interesting to see that the model with minimum number of encoder and decoder layer (i.e., 1 encoder + 1 decoder) obtains the best EAO of 0.424 in VOT2018, which outperforms the second best (i.e., 1 encoder + 2 decoders) by 1.7 points.
This result is opposite to the characteristic of Transformer architecture for object detection \cite{2020ECCV_DETR}. In our tracking network shown in \figref{figure:network}, the inputs for Transformer encoder and decoder are extracted from the shared backbone, which is the main difference from other Transformer architectures (e.g., two different language word embeddings in machine translation \cite{2017NIPS_Transformer}; image feature embeddings for encoder and candidate object embeddings for decoder in object detection \cite{2020ECCV_DETR}). We hypothesize that deep Transformer layers, which involve several times to compute encoder-decoder attention between feature embeddings from the same backbone, might make the network overfitting to the training dataset during training, thereby degrading the robustness to unknown tracking object. Therefore, we decide to design our Transformer network with 1 encoder and 1 decoder, which also leads to the advantages in inference speed and model parameter size.

\begin{table}[!t]
  \begin{center}
    \begin{tabular}{ccccccc}
    \hline
    encoder & 1  & 2 & 1  & 2 & 3 \\
    decoder & 1  & 1 & 2  & 2 & 3 \\ \hline
    EAO & 0.424 & 0.362 &  0.407 & 0.396 & 0.366  \\ \hline
    \end{tabular}
    \caption{Analysis of the impact of number of encoder and decoder layers in Transformer architecture on the offline part of \figref{figure:network}. VOT2018 dataset is used.}
    \label{table:transformer_layers}
  \end{center}
\end{table}

\begin{figure}[t]
  \begin{center}
    \includegraphics[width=\columnwidth]{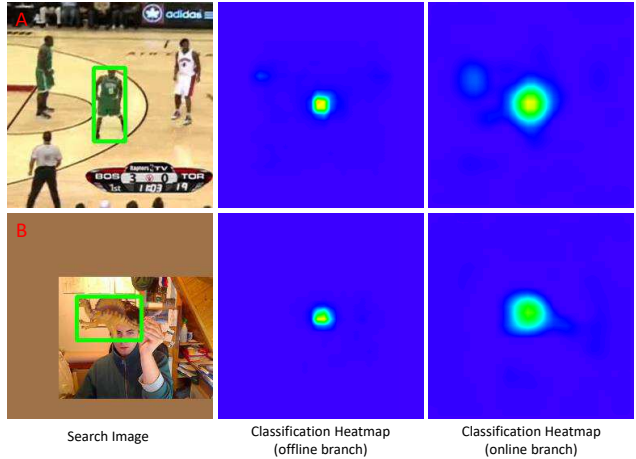}
    \caption{Visualization of classification heatmaps. The 1$^{st}$ column: search images following the cropping rule of SiamFC \cite{2016ECCVWS_SiamFC}, the 2$^{nd}$ column: classification heatmaps from the offline branch (top part of \figref{figure:network}), and the 3$^{rd}$ column: classification heatmaps from the online branch (top part of \figref{figure:network}).}
    \label{figure:heatmap}
  \end{center}
\end{figure}

\paragraph{Impact on positional encoding.}
The positional encoding shown in \figref{figure:transformer} is important to identify the spatial position of sequence in Transformer architecture. In original Transformer \cite{2017NIPS_Transformer}, the positional encoding is added to every element of the sequence. However, in case of tracking task, search image might contain an area that out of the original image such as \figref{figure:heatmap}(B). Following the cropping rule of SiamFC \cite{2016ECCVWS_SiamFC}, this area is padded by the average channel values of the whole image. We evaluate the influence of the positional embeddings on such a padding area during training and inference, and compare the performance between two cases: padding area with positional embeddings ({\bf w/ PE}) and padding area without positional embeddings ({\bf w/o PE}). As shown in \tabref{table:positional_embedding}, no positional embeddings in padding area can bring a large improvement of 4.8 points in terms of EAO compared to adding positional embeddings in padding area. We consider that, without positional embeddings in padding area, the query and key sequences $Q$, $K$ in  \equref{eq:qkv_matrices} will have  the same embedding values in padding area, which leads to the attention weight matrix $A$ in \equref{eq:qk_weights} having the same weight related to the padding area. Then, more weights are assigned to the position pairs related to the target object. Thus, such a special ``mask'' operation can help both self- and cross-attention mechanisms to emphasize the area of target object.

\begin{table}[!b]
  \begin{center}
    \begin{tabular}{ccccc}
    \hline
      &  w/ PE & w/o PE \\ \hline
     EAO & 0.376 &  0.424 \\ \hline
    \end{tabular}
    \caption{Analysis of the impact of positional embedding in padding area such as \figref{figure:heatmap}(B). Two different cases during training and inference: padding area with positional embeddings  ({\bf w/ PE}) and padding area without positional embeddings ({\bf w/o PE}). VOT2018 dataset is used.}
    \label{table:positional_embedding}
  \end{center}
\end{table}


\paragraph{Analysis of classification heatmap.}
We compare the classification heatmaps obtained from the offline branch (i.e., top part in \figref{figure:network}) and the online branch (i.e., bottom part in \figref{figure:network}) as shown in \figref{figure:heatmap}. It can be confirmed that the probability model in the heatmap of the offline branch trained with focal loss \equref{eq:classification_loss}  is sharper than that in the heatmap of the online part trained with $L^2$ error model \cite{2019CVPR_ATOM}. As discussed in  \cite{2018ICCV_FocalLoss},  focal loss can solve foreground-background class imbalance (i.e., the vast number of easy negatives) during training and improve the convergence performance of loss function. During inference, such a sharp probability model can help our tracker to discriminate the target from the distractors easier. For online update model, we still use the $L^2$ error model following \cite{2019CVPR_ATOM} to ensure the robustness against the appearance changes of target object. As shown in \tabref{table:vot}, integrating this online update module yields another improvement of 6.9 EAO points in VOT2018 benchmark, showing the scalability of the proposed framework.

\section{Conclusions}
\label{sec:conclusion}

In this paper, we have presented a unified framework to end-to-end train a network based on Transformer encoder-decoder architecture for visual tracking instead of using explicit cross-correlation operation like recent  Siamese-network-based trackers.
Our model exploits self- and encoder-decoder attention mechanisms in Transformer architecture to globally reason about the target object using pair-wise relations between template and search image features. Moreover, the proposed classification and regression heads based on shape-agnostic anchors ensure the robust tracking for target localization.
While our proposed tracker integrated with a plug-in online update model achieves competitive performance against SOTA trackers in most benchmarks, the performance on large scale datasets, TrackingNet and LaSOT, still has the room for improvement.
In the future work, we will extend our Transformer architecture by using sequential frames in temporal manner to address the challenges of fast target motion, appearance changes, and occlusion without integrating an additional online branch.


{\small
\bibliographystyle{ieee_fullname}
\bibliography{main}
}

\end{document}